%% file: powwow.tex
\documentclass{article}

\usepackage{microtype}
\usepackage{graphicx}
\usepackage{subfigure}
\usepackage{booktabs} 

\usepackage{hyperref}

\usepackage[accepted]{icml2020}
\usepackage{comment}
\usepackage{microtype}
\usepackage{multirow}

\icmltitlerunning{Submission and Formatting Instructions for ICML 2020}

\begin{document}

\twocolumn[
\icmltitle{Pow-Wow: A Dataset and Study on \\ Collaborative Communication in Pommerman}



\icmlsetsymbol{equal}{*}

\begin{icmlauthorlist}
\icmlauthor{Takuma Yoneda}{ttic}
\icmlauthor{Matthew R. Walter}{ttic}
\icmlauthor{Jason Naradowsky}{pfn}
\end{icmlauthorlist}

\icmlaffiliation{ttic}{Toyota Technological Institute at Chicago, Chicago, IL}
\icmlaffiliation{pfn}{Preferred Networks, Inc., Tokyo, Japan}

\icmlcorrespondingauthor{Takuma Yoneda}{takuma@ttic.edu}
\icmlcorrespondingauthor{Matthew R. Walter}{mwalter@ttic.edu}
\icmlcorrespondingauthor{Jason Naradowsky}{narad@preferred.jp}

\icmlkeywords{Machine Learning, ICML}

\vskip 0.3in
]



\printAffiliationsAndNotice{} 

\input{00_abstract.tex}

\input{01_intro.tex}
\input{02_pommerman.tex}

\input{03_data.tex}

\input{04_analysis.tex}
\input{05_rl.tex}

\input{06_related.tex}
\input{07_discussion.tex}
\input{075_acknowledgements.tex}

\bibliography{powwow}
\bibliographystyle{icml2020}

\appendix
\input{08_appendix}

\end{document}

%% file: 00_abstract.tex
\begin{abstract}
In multi-agent learning, agents must coordinate with each other in order to succeed.  For humans, this coordination is typically accomplished through the use of language.  In this work we perform a controlled study of human language use in a competitive team-based game, and search for useful lessons for structuring communication protocol between autonomous agents.


We construct \emph{Pow-Wow}, a new dataset for studying situated goal-directed human communication.  Using the \textit{Pommerman} game environment, we enlisted teams of humans to play against teams of AI agents, recording their observations, actions, and communications. 
We analyze the types of communications which result in effective game strategies, annotate them accordingly, and present corpus-level statistical analysis of how trends in communications affect game outcomes.  Based on this analysis, we design a communication policy for learning agents, and show that agents which utilize communication achieve higher win-rates against baseline systems than those which do not.
\end{abstract}


%% file: 01_intro.tex
\section{Introduction}
Collaboration is an integral part of human society, and an important factor to our evolutionary success.  Through the development of language, communication become possible across great distances for the first time. 
But despite progress in multi-agent learning, language has not yet been incorporated effectively into this paradigm.

A natural way to incorporate language into multi-agent learning is to learn both the communication protocol and the game strategy together, from the ground up.  In theory, training to minimize the task loss may then allow agents to learn an optimal communication strategy.  Several examples of prior work have achieved various degrees of success (under the title of emergent communication, EC), but because agents begin from a blank slate, current EC work focuses only on the simplest of environments, and only the simplest of communications (often a binary decision, \cite{zhang2013coordinating, evtimova2017emergent, eccles:biases}).








We approach the problem from the opposite direction: human players already know language, and are capable of conveying concepts they believe are important for gameplay.  In  a collaborative game setting, what concepts do such players choose to communicate, and how do they choose to express it?  What lessons can we learn from studying how humans actually achieve collaboration in such games, and can we transfer this knowledge into agents' policies?

\begin{figure}[tp]
    \centering
    \includegraphics[width=0.80\linewidth]{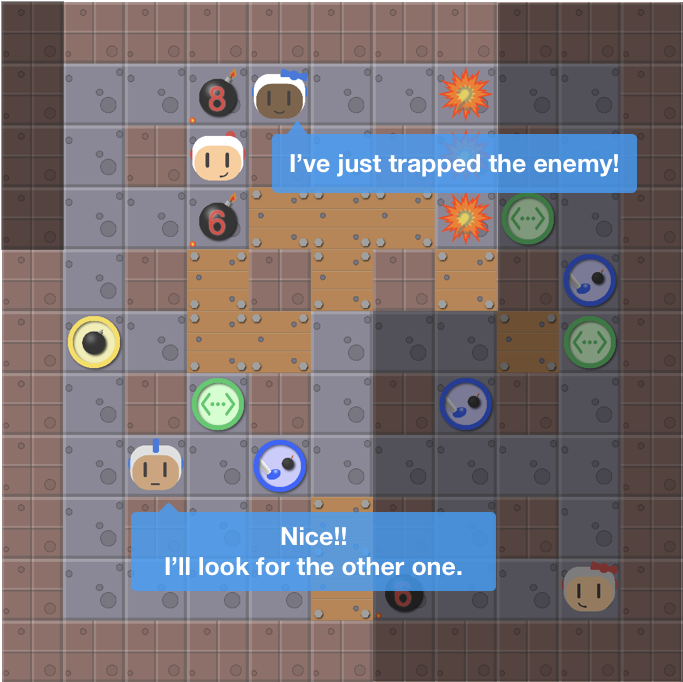}
    \caption{Cooperative play in Pommerman. Brighter area is the union of observable areas by agents that have a blue ribbon.
    }
    \label{fig:pom-image}
\vspace{-2em}
\end{figure}

In this work we study communication in the team-based competitive game environment of \emph{Pommerman}~\citep{pommerman}.  The world of Pommerman is a partially-observable map where each agent has only a limited view of the world around them (Fig.~\ref{fig:pom-image}).  Recent Pommerman shared tasks have emphasized the potential role of language in the task (\emph{Team Radio}\footnote{https://www.pommerman.com/competitions}), allowing agents to send coded messages to their teammates via a communication channel.
While communication is not strictly necessary for success in Pommerman, each agent will often have information their teammate does not due to partial observability, and agents who communicate can gain a strategic advantage.

We present \emph{Pow-Wow}, a dataset of human actions,  and communications in Pommerman.  We enlisted participants to play in 2-human vs. 2-AI matches, and recorded the game state, the player actions, and text-based communication between players.
After data collection, we examine the ways in which teammates communicated to one another, and construct an ontology of communicated concepts.  Using this ontology we manually annotated each communication in the dataset.  Using statistics of effective communication in the recorded games as a springboard, we train reinforcement learning (RL) agents to employ similar communication strategies. Even though our communications are simple, we find our augmented agent achieves higher win-rates than non-communicative agents.

We release the Pow-Wow dataset and data-collection code for further research and present this study as a workflow for building communicative game agents.








%% file: 02_pommerman.tex
\section{Pommerman}
\label{sec:pom}


Pommerman \citep{pommerman} is a multi-agent game environment based on the classic console game, Bomberman. 
In Pommerman, four players play in two-person teams, with each player initially placed in a separate, randomly-chosen corner of the map.  The game concludes when all members of a team have been eliminated.

At each step, players can choose to either stay, move to a neighboring cell (cardinal: north/south/east/west), or plant a bomb on the spot.  Bombs explode in the same cardinal directions, and have a set timer, allowing time for players to escape the blast.  Bombs have two purposes: (1) to destroy blocks, which can create new paths on the map and reveal power-ups, and (2) to kill enemy agents (or themselves!).  Additional game mechanics are described in Appendix~\ref{app:game-mechanics}.

%% file: 03_data.tex
\section{The Pow-Wow Dataset}

There are several official variants of Pommerman.  
The third competition, entitled ``\emph{Team Radio}'' 
allows teammates to send each other a pair of integer value ``messages''.  
While these constraints 
are too strict to allow for human-like communication, we explore what types of communication humans are likely to use, as this may provide insight into how to best utilize the channel.  For this we modified the original Pommerman environment, developed an annotation application, and collected data from human trials.

\vspace{-8px}
\paragraph{Modifications to Pommerman}
To ensure that human players can communicate sorely via text-based chat, we prepared separate terminals for  those who play the game together. We added a chat window on each terminal, making sure that the messages are synchronized over the network with the other terminal and the game state. At each turn, humans are able to send an unbounded number of messages to communicate with the teammate before taking an action.


In addition, blocks are placed randomly in Pommerman, which can sometimes isolate agents and slow down the pace of the game.
We adopt the Bomberman-style placement of indestructible blocks (visualized as brown blocks in Figure \ref{fig:pom-image}) in cells where $(i,j)$ describe a coordinate system and $i$ and $j$ are both even numbers.  This makes it generally more difficult to escape bomb blasts, and gives additional strategic value to bomb chains.  Other detailed modifications are summarized in Appendix \ref{app:mods-to-pommerman}.


\subsection{Annotation Environment}

We adopt a 2 vs 2 team match setting, where one team consists of human participants, and the other is the top-ranked system from a previous competition, {\it hakozakijunctions}~\citep{hakozaki}.  This system uses Monte Carlo tree search 
and is especially good defensively, where look-ahead search allows it to identify and avoid potential traps.  However, these agents cannot communicate with one another, and can be defeated through coordinated attacks.

We implemented a text-based chat window for players to communicate, and also altered the game graphics to annotate bombs with their blast range and remaining time, which are otherwise provided to computer agents but not shown graphically in the original version.
We release code for our data-collection framework, including these modifications, as well as the collected data\footnote{https://takuma-ynd.github.io/pow-wow/}.

\subsection{Data Collection}


We enlisted student participants to play the game, each seated at a separate terminal.  For new players (the most common scenario) we provide a video tutorial which introduces the rules and how to play the game.  We adopted the following additional rules: (1) players cannot communicate verbally (i.e., only text-based communication is allowed) ; and (2) if any human player is eliminated, the game is terminated 
(as further communication is not possible).


We collected 90 matches 
from 60 people with 2,832 messages in total.  After filtering out the matches that did not have any communication or ended too early, we were left with 80 games by 59 people, with 2,513 messages in total.  The average game length was between 15 and 20 minutes.


Playing against the competitive enemy agent ({\it hakozakijunctions}), human teams won 15 times (18.75 \%), tied 17 times (21.25\%), lost 48 times (60.00 \%) out of 80 matches. 






%% file: 04_analysis.tex

\section{Identifying Effective Communication}
Having collected the data, we now turn to understanding it.  Based on preliminary analysis of the communications observed in the game, we constructed an ontology for classifying each \textit{dialogue} as belonging to one or more of 25 hierarchical categories
(the full depiction of this ontology and example messages are provided in Appendix~\ref{app:ontology}).  Note that \textit{dialogue} here refers to a sequence of messages exchanged between teammates in one timestep.
We annotate each dialogue rather than individual messages 
, as we observed that the main idea of a conversation is sometimes conveyed over the course of many individual messages.



Inter annotator agreement is calculated by a variant of Krippendorff’s Alpha \citep{iaa_metric}, obtaining $\alpha = 0.772$. The frequencies of communication types over time are summarized in Appendix~\ref{app:communication-over-time}.

\begin{figure}[t!]
  \includegraphics[width=0.97\linewidth]{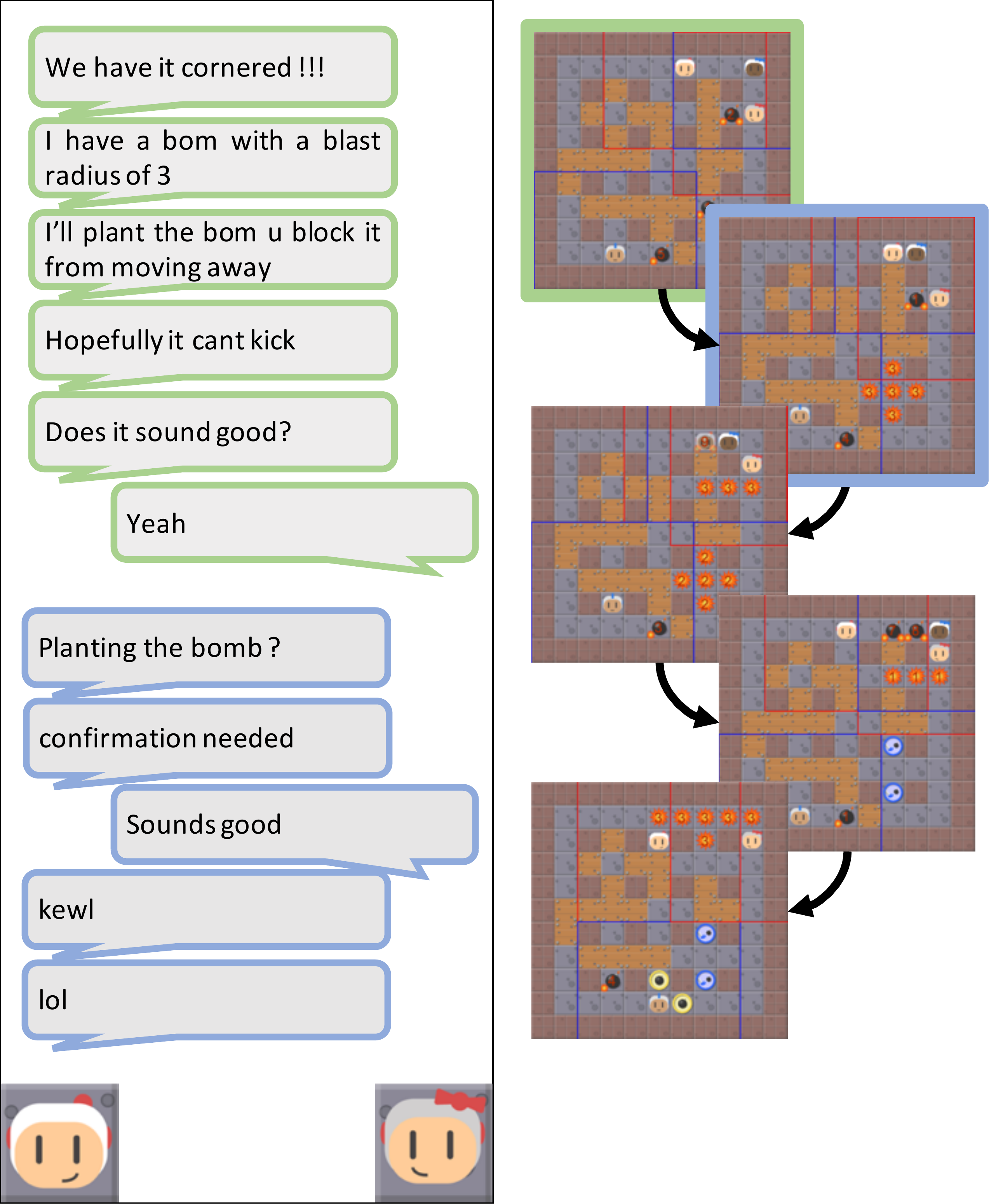}
  \caption{Action sharing in a collaborative play. Blue and red square contours on the board shows each agent's visible region.}
  \label{fig:collab_play}
  \vspace{-2em}
\end{figure}

\vspace{-0.8em}
\paragraph{Common Strategies}
We observe that regardless of match outcome, players frequently coordinate at the beginning of the game to identify the target enemy and their locations (\textit{target enemy, own location (area)}).
An example of this communication is ``\emph{Let's corner top right enemy together}''.  This initial coordination establishes a high-level game strategy. Players typically converge to the location of the first enemy sighted, presumably to gain a 2-on-1 advantage over them.  This is common across all outcomes, but moreso in wins than ties, and in ties than losses.
\vspace{-0.8em}
\paragraph{Effective Communication}
\label{sec:effective-comm}
In successful games, we observe players prefer an action-centric communication style, often suggesting actions to the teammate (\textit{action suggestion} and \textit{own action}).  Action heatmaps in Appendix \ref{app:communication-over-time} reveal how important such communications are in coordinating attacks at end game.  In comparison, tie games contained more late game stage communications describing observations, and more descriptions of their own behavior.  
It appears that in tie games players' coordination at end game is more ambiguous: ``if I describe my intention, maybe my teammate will know what to do''. Interestingly, the more effective strategy seems to be simply directing the other agent, which coincides with common emergent communication tasks. One of such examples are shown in Figure \ref{fig:collab_play}.

%% file: 05_rl.tex
\section{Learning a collaborative play}
\label{sec:rl}

As we have shown in the preceding section, a study of human behavior and communication in multi-agent games can yield interesting insights into effective game strategies, but can these lessons be effectively transferred into game playing agents?  We explore this question by training agents with RL to utilize some of the communication strategies we find in human-play data, and evaluate their performance with respect to baseline non-communicative agents.



\subsection{Training agents with communication}
As we have described in Section \ref{sec:effective-comm}, we find that communicating a target enemy in the early game can result in more successful game outcomes.  We also find its location and agent's own location to be commonly shared and feasible for learning agents. Based on this, we design a communication strategy to share these information.



We simulated this communication strategy in a simple manner. For every time step, the location of an agent is directly added in its teammate agent's observation. Whenever an agent observes the target enemy, we hallucinate the enemy in the teammate agent's observation, preserving its location.  Note that in this approach, the agent receiving the message does not need to ``interpret'' it as is done in EC research; it is simply represented as part of the ordinary observation.

We used asynchronous advantage actor-critic (A3C) \citep{a3c} to train the policy.
Due to the challenging multi-agent credit assignment problem, training RL agents  purely from the sparse task reward can be difficult~\citep{skynet, combat}.  We lessen this difficulty by employing a curriculum learning approach~ \citep{skynet}. 
Details on training are summarized in Appendix~\ref{app:learning-collab-agents}.

\subsection{Results}
%

We evaluate our agent policies by measuring their performance against the heuristic Pommerman baseline system, (\textit{simple~agent}). First we train an agent against random agents to learn basic strategy (\textit{vs-random}).  We then fine-tune this system by training it against the more competitive \textit{simple agent}.  We fine-tune one model with communication, and one without.
We then run a competition pitting \textit{simple agent} against each of our systems, and record the win-loss ratio over 50 matches for each scenario.  Results are shown in Table~\ref{tb:rl-result}.


We find agents trained with communication significantly outperform ones without it. 
There seem to be two factors which lead to this competitive advantage.  First, is simply that communicating the location of the first enemy seen causes agents to coordinate against it, rather than dividing their efforts into whichever enemy each encounters first.  This ``high-level'' strategy is one of the main factors we observe in successful human play, and we can rule out our communication having any effect on close-quarters 2-on-1 play, since agents which are close to each other can naturally observe most of each other's around without any need for communication.
However, a second benefit is that it appears that communication allows agents to encounter 2-on-1 situations more frequently in training.  More 2-on-1 situations in training results in more opportunities to learn effective close-quarters end-game style play.

\begin{table}[t]
\centering
\begin{tabular}{l rrr}
\toprule
& win & tie & lose \\
\midrule
vs-random      & 2   & 17  & 31 \\
fine-tuned w/o comm.      & 8   & 38  & 4  \\
fine-tuned w/ ~~comm. & 21  & 22  & 7    \\
\bottomrule
\end{tabular}
\caption{Game results (50 games each) of each agents against \textit{simple-agent}}
\label{tb:rl-result}
\end{table}

%% file: 06_related.tex
\section{Related Works}

Finding effective natural language communication is an important topic in our work.
~\citet{conversational-marker} 
identify linguistics cues that correlate highly with effective communication.  They find that markers present very early in communication are good indicators of ultimate success, in the same way we find that
sharing a target (\emph{target enemy})
at the start of the game
often result in wins, as opposed to more self-concerned communications in ties (\emph{I'm in the bottom left corner}), and irrelevant communication in losses.

In line with dataset construction by observing interaction between humans and their communication, several works presented dataset on a collaborative \citep{nonverbal-behavior-to-identify-leaders, who-will-get-the-grant} or competitive task \citep{idiap-wolf-corpus}. However, their focus is on non-verbal communication with multi-modal data, 
in a less controlled environment compared to ours. 

%% file: 07_discussion.tex
\section{Discussion}
In this work we present not only a new dataset for the study of human collaborative communication, but a proof of concept of a 
linguistically-motivated approach to designing cooperative multi-agent systems.  Even though our dataset is small by modern crowd-sourced standards, working with a limited budget we were able to collect and analyze the data to yield valuable insights.  We prove the usefulness of these insights by structuring the design of a learning agent around the human behaviors which were salient in the data, and showed that the resulting policies were superior against baseline agents.

A natural criticism may be that the communication protocol we adapt from Pow-Wow is simple.  
But Pow-Wow also contains more nuanced instances of collaboration, and learning strategy from a mix of limited observations and prior knowledge is an important challenge for the community.  For instance, using recent breakthroughs in large pre-trained language models, the diversity of messages may be effectively extended through paraphrasing.




But we contend that datasets like Pow-Wow, which combine human communication, observation, and action, all temporally aligned, present new research opportunities.  While in this work we focus on developing a communication protocol, it is also possible to try to learn a mapping between communication and future actions to understand how humans express abstract strategy (``\emph{Do a pincer attack!}'').  In a similar way, this data can also be used to train agents which react reasonably to a human teammate's speech commands.  

One can also look solely at the human observations and actions, to learn more human behavior.  For instance, it was apparent in the previous year's competition that bomb chains, despite being a fundamental aspect of the game's dynamics and a common human tactic, were not being utilized by the participating systems -- both for agents trained from data and those which are mostly search-based.  We hope that future work can utilize our data to improve agent strategy appropriately.

Moving beyond Pommerman, there are great opportunities for large-scale studies of the sort we perform here.  In Pommerman, we must enlist participants to play, and where data gathering and annotation is time-consuming and costly.  But in the world of online gaming, where gamers constantly utilize spoken language for coordination, the ability to collect such data is nearly limitless.  We hope this work can serve as motivation for the construction of such a dataset.

%% file: 075_acknowledgements.tex
\section*{Acknowledgements}
We would like to express our deep gratitude to the folks in TTIC and University of Chicago who participated in the study, and the brilliant people in Preferred Networks and colleagues in our robotics lab for providing us very informative advice. 

We would also like to give a special thanks to Takeru Oba, Takahiro Maeda and Takeshi Onishi who provided a lot of insightful observations during the study.

%% file: 08_appendix.tex
\clearpage
\section{Supplemental Material}

\subsection{Game Mechanics}
\label{app:game-mechanics}
Here we provide an overview of game mechanics.
The game map consists of 11 $\times$ 11 cells as shown in Figure \ref{fig:pom-image}, and each agent begins from one of the four corners: $(1, 1), (1, 9), (9,1)$ or $(9,9)$, where indexing starts from zero.

\paragraph{Power-Ups}
The presence of power-ups adds an interesting gameplay dynamic, as agents may have disproportionate abilities when they encounter each other. There are three kinds of power-ups:  
\begin{itemize}
    \item Ammo, increases the number of bombs a player can have on the board simultaneously.
    \item Blast range, increases the number of tiles in each direction that are touched by an exploding bomb.
    \item Kick, gives the player the ability to push a bomb forward, sending it across the map until either its timer reaches zero and it explodes, or until it collides with an object.
\end{itemize}

\paragraph{Bomb Chaining}
When a bomb explodes, any bomb which comes into contact with the blast will also explode.  This creates the potential for chain reactions which become difficult to calculate or plan for.  While a single, isolated bomb blast is easy to avoid, carefully constructed bomb chains, especially those created by two players interacting, can be very destructive.  This is a key dynamic in playing Pommerman effectively.

\paragraph{Action-Cancellation}  
In Pommerman, all agents' moves are executed simultaneously.  This creates the potential for conflict when two players attempt to move into the same unoccupied space.  The game resolves this scenario by cancelling both moves, and agents remain in their original positions. While this is merely an artifact of the game engine, we observe both human and computer players exploited it to trap other players where technically there is an escape route, but repeated cancellations prevent an agent from moving and escaping a bomb blast.

\subsection{Modifications to Pommerman}
\label{app:mods-to-pommerman}
We add several modifications to original Pommerman in its game map, definision of win/loss/tie, and game engine and interface.
\subsubsection{Modifications to the game map}
In Pommerman, indestructible blocks are placed randomly. Random placement of the blocks makes it easier for agents to avoid the explosions, exceedingly increasing the time of the match. In preliminary test matches, we found that some games exceeded 300 time steps, equating to more than two hours per game.

To mitigate this, we placed the blocks in cells $(i, j)$ where $i$ and $j$ are even numbers, and also filled the outer-most edges with the blocks from the beginning. In addition, we decided to utilize
wall-collapsing, a feature from Bomberman that is
not utilized in Pommerman competition. If games
exceed a certain number of time steps ($t$ = 100 in
our trials), the outer edge of the map is incrementally
filled in with additional indestructible blocks. This has the
effect of reducing the usable map space, forcing
agents into closer proximity, and ultimately forcing
the end of the game.
After introducing these changes a typical match duration
was approximately 15 minutes, and no match
exceeded 200 steps.

\subsubsection{Altered definition of game outcome}
Now, we describe the definitions of game results (win, tie, lose) in this data collection. Since we terminate the game when any human player is eliminated, there are some cases that is ambiguous which team wins the game. Throughout the paper except Section 5, we define win as eliminating all the AI agents, tie as either all agents dying at the same time or one of the AI agents being eliminated
when game is terminated, and lose as all AI agents being alive.

\subsubsection{Text-based chat window}
Another modification to the system is the addition of a chat window. The chat functionality is implemented as a separate local client-server where one machine (laptop) acts as both one instance of the game and chat window, but is also acting as a server for a second instance running on a remote machine. At each time step players are allowed to participate in an open-ended dialogue in the chat window prior to taking an action.  We considered only allowing players to send a single message at each time step to more closely resemble the Pommerman radio challenge rules, but found that this encouraged players to resolve complicated discussions over multiple time steps, making it difficult to analyze.

\subsection{Details on learning communicative agents}
\label{app:learning-collab-agents}
The default observation for an agent in Pommerman contains partially observable board, existing bombs’ life and blast strength, how long any flame will continue to persist on the board, player’s ability (ammo, blast strength, kick), list of alive agents including enemies, and time steps. We encoded observations that contains spatial information such as board, bombs, flames, and target enemy position (if applicable), into a $6 \times 11 \times 11$ tensor. And the rest of observations into a 17 dimensional vector.

In addition to these, we calculated the board state in the next step according to the game engine, assuming all agents will choose stop action (stay on the spot), in order to obtain a reduced list of action suggestion (implemented by \citet{skynet}).  This filters the list of actions that would result in immediately death (such as jumping into flames, standing in the path of a flame, cornering oneself etc.). For the usage of action suggestion, it should be noted that we just included it as an additional observation without filtering actions based on the suggestion, as is done in \citep{skynet}. We find that this is an important factor in learning a policy effectively. By increasing the survival rates of agents, we improve their chances of receiving end-game reward and shaping other behaviors.

In terms of the agent model, to construct representations of the observed game state we use a 3-layer CNN of size (8, 16, 16), and kernel size 3. The output of this network is flattened and concatenated with the 17 dimensional communication vector. The policy network is a 2-layer MLP of
each output dimension being 128 and 64, which internally concatenates the action suggestion n-hot vector (6 dimension). This output is separately fed to softmax layer to obtain action probabilities and linear layer to obtain value. All the layers except the final layer described above is followed by rectified linear units (ReLU).

\subsubsection{Curriculum learning}
We adopt a curriculum learning approach, similar to \citep{skynet}. Each phase in the curriculum is described below.

\begin{itemize}
    \item \textbf{Phase 1:}
In the first stage, we trained the agent against static agent which never move or place a bomb, with positive reward for destroying woods, picking up items and negative reward for dying.
During this period, agent learned basic skills such as placing bombs, avoiding explosions and collecting items.
\item \textbf{Phase 2:} After 8.5 million steps, the static agents are replaced by \textit{random-without-bomb} agents which moves around randomly but never place a bomb. At this point we replaced the previous positive reward with seeing enemy in sight, being close to an enemy (based on the shortest path) and killing an enemy. Starting from this phase, the latest training policy was used as learning agent's teammate.
\item \textbf{Phase 3:} By 5.2 million steps, the trained agent can play competitively against \textit{random-without-bomb} agent.
However, we note that trained agents sometimes accidentally kill themselves with their own bombs, which can create a loss even when playing against \textit{random-without-bomb} agents. We then replace the opponents with a scripted agent ({\it simple-agent}) which is provided by default in Pommerman environment.  
Starting from this point, we trained agents in two different settings: one that has communication, and the other that does not.
\end{itemize}

\onecolumn
\section{Ontology}
\label{app:ontology}
\begin{figure*}[h]
  \includegraphics[width=\textwidth]{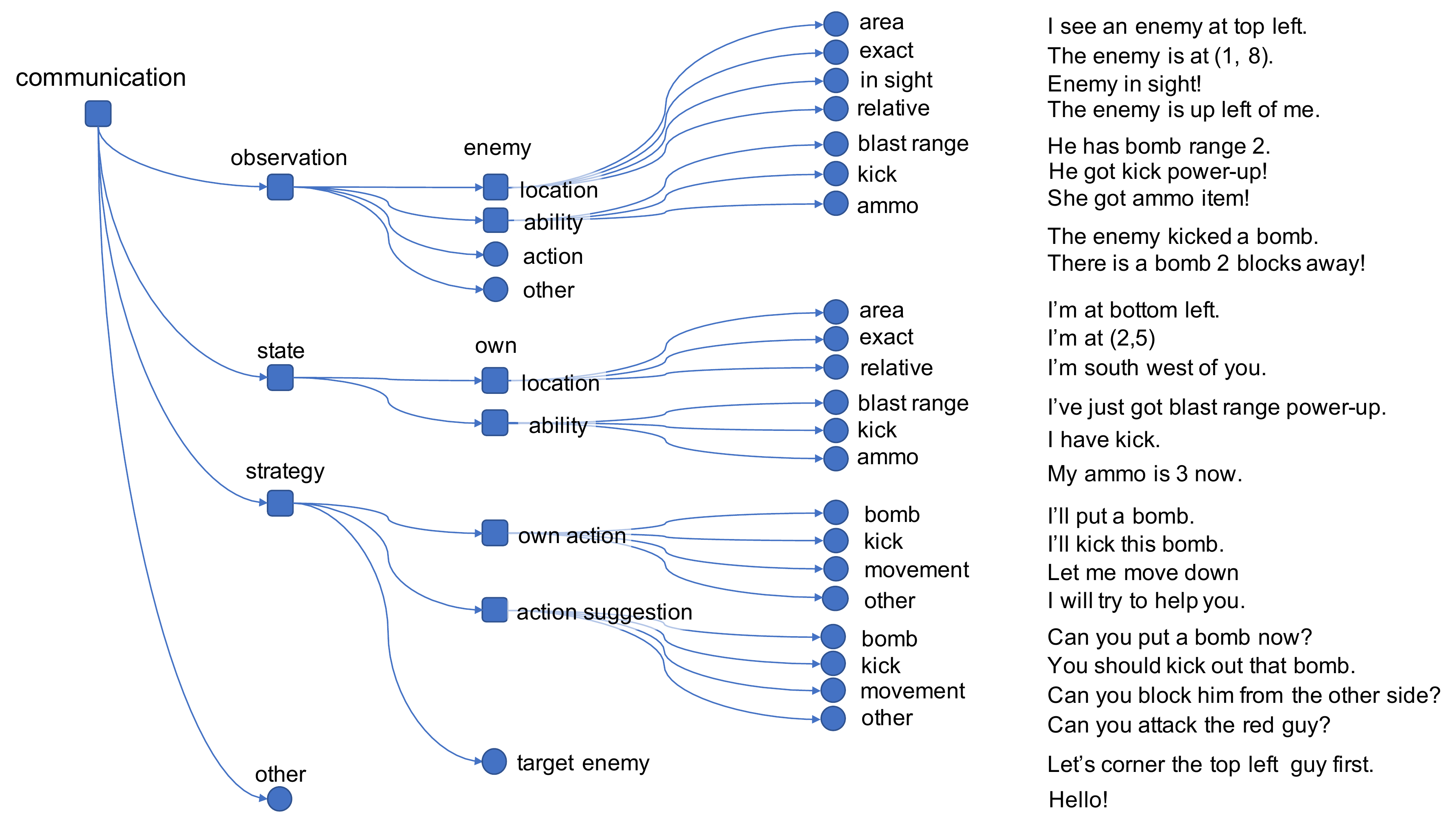}
  \caption{Pow-Wow Pommerman communication ontology.}
  \label{fig:ontology}
\end{figure*}

\clearpage
\section{Communication over Time}
\label{app:communication-over-time}
\begin{figure*}[h]
  \includegraphics[width=\textwidth]{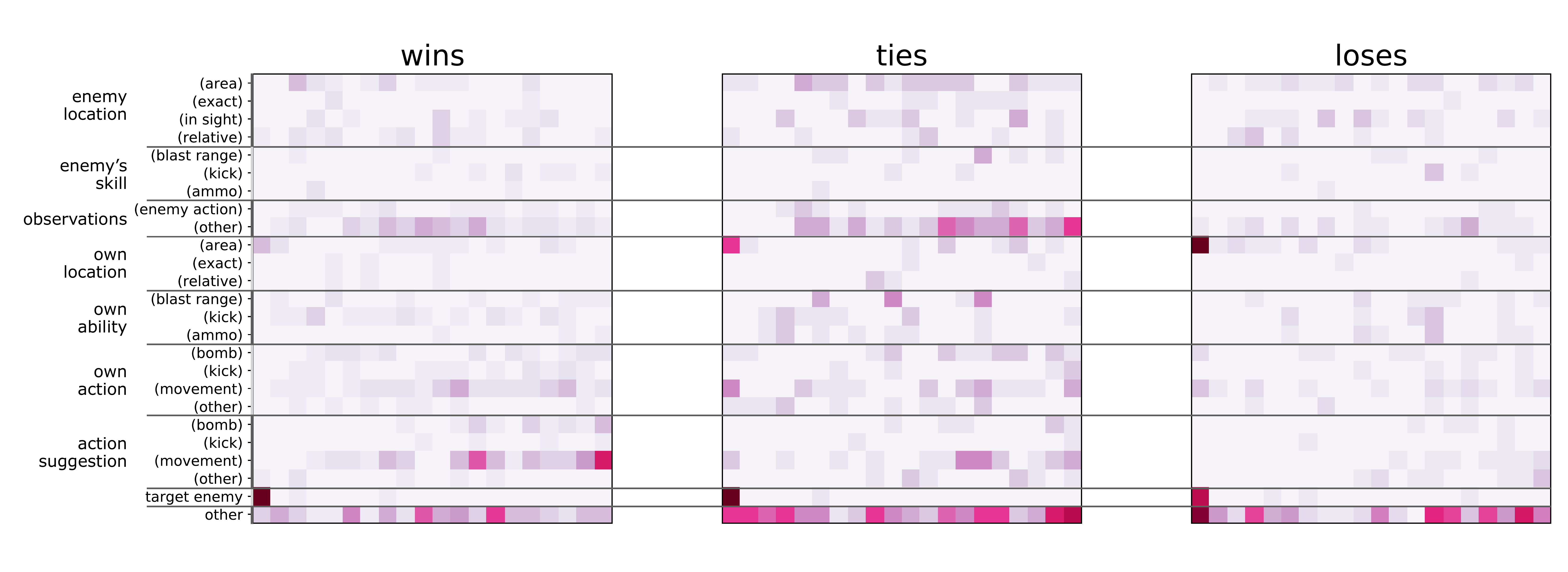}
  \caption{Types of messages across timesteps. The x-axis is a time step normalized by the length of the corresponding game. Some abbreviations: \emph{loc} refers to location.  Winning games exhibit more action suggestions toward end-game than ties or losses.  In losing games, communication is used unproductively (\emph{other}).}
  \label{fig:messages_over_time}
\end{figure*}

